\documentclass[11pt,a4paper]{article}
\usepackage[dvipsnames]{xcolor}
\usepackage[hyperref]{naaclhlt2019}

\usepackage{times}  %Required
\usepackage{helvet}  %Required
\usepackage{courier}  %Required
\usepackage{url}  %Required
\usepackage{graphicx}  %Required

\usepackage{latexsym}
\usepackage{multirow}
\usepackage{amssymb}
\usepackage{mathtools}
\usepackage{booktabs}
\usepackage{float}
\usepackage{enumerate}
\usepackage{natbib}
\usepackage{amsmath}
\usepackage{caption}
\usepackage{adjustbox}
\usepackage{url}

\DeclareMathOperator*{\argmax}{arg\,max}

\newcommand{\specialcell}[2][c]{%
  \begin{tabular}[#1]{@{}l@{}}#2\end{tabular}}
  
\newcommand{\xdownarrow}[1]{%
  {\left\downarrow\vbox to #1{}\right.\kern-\nulldelimiterspace}
}

\aclfinalcopy

\begin{document}

\title{GraphIE: A Graph-Based Framework for Information Extraction}
\author{Yujie Qian$^1$, Enrico Santus$^1$, Zhijing Jin$^2$, Jiang Guo$^1$, and Regina Barzilay$^1$ \\
$^1$Computer Science and Artificial Intelligence Laboratory, MIT \\
$^2$Department of Computer Science, The University of Hong Kong \\
\url{{yujieq,jiang_guo,regina}@csail.mit.edu, {esantus,zhijing}@mit.edu}
}
\maketitle
\begin{abstract}

% RB: most of the XXX focus on modeling of local dependencies, xxx non-local is valuable
% 4 extraction tasks/domains -> to be consistent
%%%CHANGED
Most modern Information Extraction (IE) systems are implemented as sequential taggers and only model local dependencies. Non-local and non-sequential context is, however, a valuable source of information to improve predictions. In this paper, we introduce GraphIE, a framework that operates over a graph representing a broad set of dependencies between textual units (i.e. words or sentences). The algorithm propagates information between connected nodes through graph convolutions, generating a richer representation that can be exploited to improve word-level predictions. Evaluation on three different tasks --- namely textual, social media and visual information extraction --- shows that GraphIE consistently outperforms the state-of-the-art sequence tagging model by a significant margin.\footnote{Our code and data are available at \url{https://github.com/thomas0809/GraphIE}.} 
\end{abstract}

\section{Introduction}

\begin{figure}[t!]
\centering
\includegraphics[width=.48\textwidth]{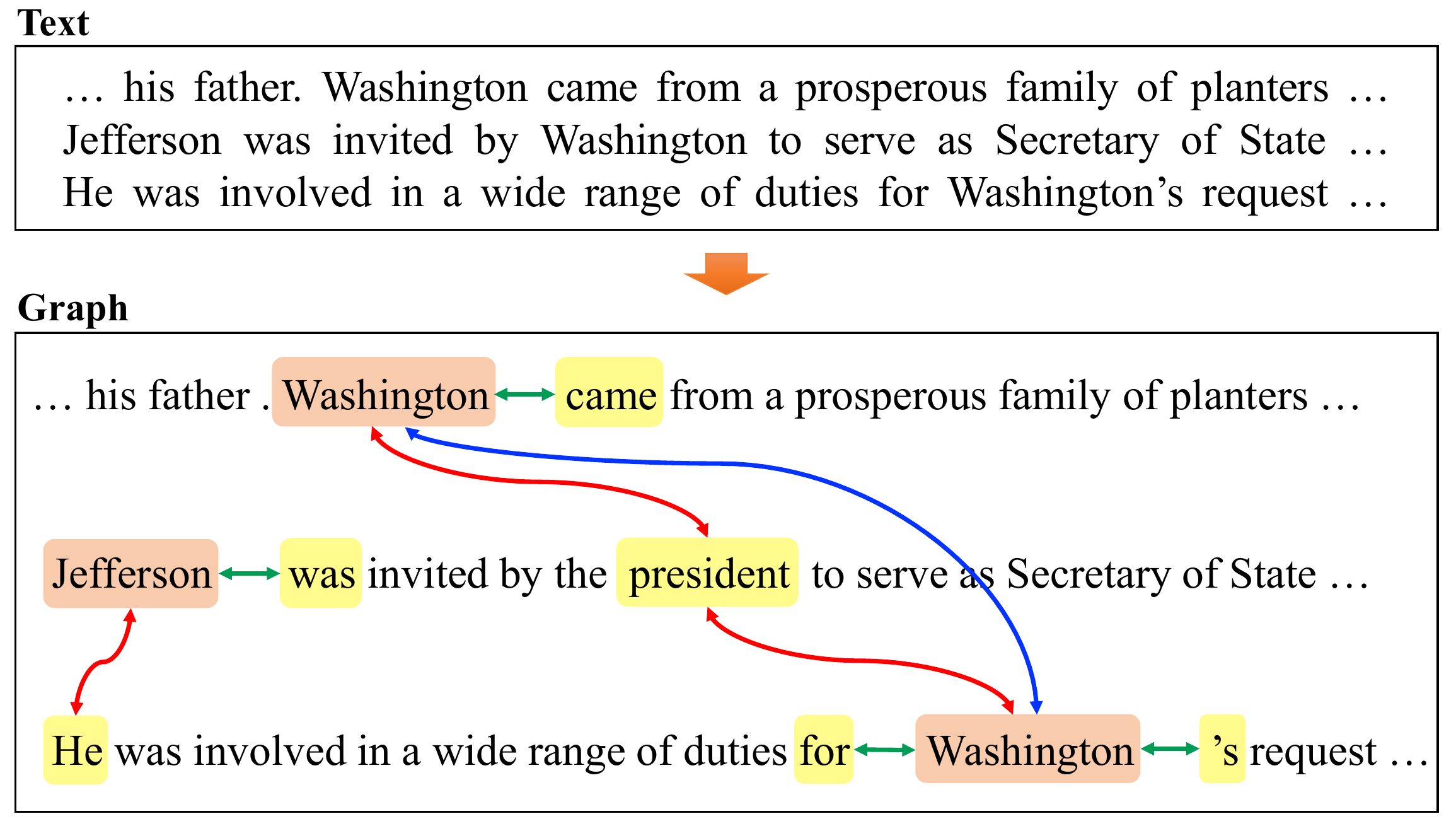}
%%%CHANGED
\caption{Example of the entity extraction task with an ambiguous entity mention (i.e.  ``...for Washington's request..."). Aside from the sentential forward and backward edges (\textcolor{ForestGreen}{green}) which aggregate local contextual information, non-local relations --- such as the co-referent edges (\textcolor{red}{red}) and the identical-mention edges (\textcolor{blue}{blue}) --- provide additional valuable information to reduce tagging ambiguity (i.e. \textsc{Person}, \textsc{Organization} or \textsc{Location}).}
% \vspace{-0.08in}
\label{fig:example}
\end{figure}

% IE can benefit from global information. 
Most modern Information Extraction (IE) systems are implemented as sequential taggers. While such models effectively capture relations in the local context, they have limited capability of exploiting non-local and non-sequential dependencies. In many applications, however, such dependencies can greatly reduce tagging ambiguity, thereby improving overall extraction performance.
% For instance, in social media IE tasks, a common hypothesize is that users that follow each other are likely to have similar attributes such as education backgrounds and jobs. Therefore, the social network structure provide additional relational inductive bias for user information extraction (i.e. user profiling) (Figure~\ref{fig:mockup}).
For instance, when extracting entities from a document, various types of non-local contextual information such as co-references and identical mentions may provide valuable cues. %will be highly beneficial. 
See for example Figure \ref{fig:example}, in which the non-local relations are crucial to discriminate the entity type of the second mention of \textit{Washington} (i.e. \textsc{Person}, \textsc{Organization} or \textsc{Location}).
%Following such assumption, we can see how non-local information becomes crucial to discriminate the entity type of the first mention of \textit{Washington}, which would instead be ambiguous if taken in isolation (Figure \ref{fig:example}). 

%[COMPLIMENTARY]

%%%CHANGED
Most of the prior work looking at the non-local dependencies incorporates them by constraining the output space in a structured prediction framework \citep{finkel2005incorporating,reichart2012multi,hu2016harnessing}.
%By design, these methods operate in the output space, modeling the structural correlation between multiple predictions. % These... can be instead captured by modeling dependencies in the input space, which can be valuable in prediction models, e.g., the co-referent dependencies in the entity extraction task as shown in Figure \ref{fig:example}.
Such approaches, however, mostly overlook the richer set of structural relations in the input space. With reference to the example in Figure \ref{fig:example}, the co-referent dependencies would not be readily exploited by simply constraining the output space, as they would not necessarily be labeled as entities (e.g. pronouns).
%For instance, the co-referent dependencies shown in Figure \ref{fig:example} cannot be readily exploited for entity extraction by constraining the output space, since the co-referent mentions are not necessarily labeled as entities. %These LIMITATION can be instead captured by modeling dependencies in the input space.
% For instance, in the task of user profiling, a user is often neighboring with multiple users and they do not necessarily have the same attribute (e.g., gender). In this case, 
% Moreover, when the constraints in the label space cannot be explicitly defined, e.g., , these approaches cannot be directly applied.
%%%CHANGED
In the attempt to capture non-local dependencies in the input space, alternative approaches define a graph that outlines the input structure and engineer features to describe it \citep{quirk2017distant}.
% One way to capture such dependencies is through defining features describing the graph structure of the input, as proposed by \cite{quirk2017distant}. 
Designing effective features is however challenging, arbitrary and time consuming, especially when the underlying structure is complex. Moreover, these approaches have limited capacity of capturing node interactions informed by the graph structure.

%%%CHANGED: WE WANT TO SHOW HOW WE ADDRESSED THE PREVIOUSLY MENTIONED LIMITATIONS.
In this paper, we propose GraphIE, a framework that improves predictions by automatically learning the interactions between local and non-local dependencies in the input space. Our approach integrates a graph module with the encoder-decoder architecture for sequence tagging. The algorithm operates over a graph, where nodes correspond to textual units (i.e. words or sentences) and edges describe their relations. At the core of our model, a recurrent neural network sequentially encodes local contextual representations and then the graph module iteratively propagates information between neighboring nodes using graph convolutions \citep{kipf2016semi}. The learned representations are finally projected back to a recurrent decoder to support tagging at the word level.

%%%CHANGED
%The adaptation of this architecture to information extraction requires to answer to two major questions: 1) which graph topology effectively captures the input underlying structure; 2) how to model the interaction between non-local contextual information learned by the graph module and the local tagging modules.

%To adapt this architecture for information extraction, we need to answer two questions. The first question relates to the design of graph topology that can effectively encode input dependencies relevant to the target task. The second question concerns modeling the interaction between the non-local contextual information learned by the graph module and the local tagging modules.

%To address these questions, we introduce a framework named GraphIE. It operates over a graph, where nodes correspond to textual units (i.e. words or sentences) and edges encode their relations, which describe task-specific structural constraints. The algorithm iteratively propagates information between neighboring nodes using a graph convolutional network, thereby constructing the non-local context representation. This contextual information is then projected back to the nodes, supporting tagging at the word level.
%. Specifically, the algorithm starts with the node representation derived by the encoder. Next, it computes graph convolutions between connected nodes, constructing the non-local context representation for the input. During decoding, this contextual information is projected back to the nodes, supporting tagging at the word level.

%%%CHANGED
We evaluate GraphIE on three IE tasks, namely textual, social media, and visual \citep{Aumann:2006:VIE:1147970.1147976} information extraction. For each task, we provide in input a simple task-specific graph, which defines the data structure without access to any major processing or external resources. Our model is expected to learn from the relevant dependencies to identify and extract the appropriate information.
% \citep{tjong2003introduction,krallinger2015chemdner}
% We also validate its performance in a novel visual information extraction task, where the spatial layout can be used to infer valuable information. %In this task we demonstrate that our framework performs well also on unseen layouts.
% In the social media IE task, we aim at extracting users' \textsc{education} and \textsc{job} from their tweets, while in textual IE, we target at a discourse-level named entity recognition task.
Experimental results on multiple benchmark datasets show that GraphIE consistently outperforms a strong and commonly adopted sequential model (SeqIE, i.e. a bi-directional long-short term memory (BiLSTM) followed by a conditional random fields (CRF) module). 
Specifically, in the textual IE task, we obtain an improvement of $0.5\%$ over SeqIE on the \textsc{CoNLL03} dataset, and an improvement of $1.4\%$ on the chemical entity extraction \citep{krallinger2015chemdner}. In the social media IE task, GraphIE improves over SeqIE by $3.7\%$ in extracting the \textsc{Education} attribute from twitter users. In visual IE, finally, we outperform the baseline by $1.2\%$.

\section{Related Work}
\label{related_work}

The problem of incorporating non-local and non-sequential context to improve information extraction has been extensively studied in the literature. The majority of methods have focused on enforcing constraints in the output space during inference, through various mechanisms such as posterior regularization or generalized expectations \citep{finkel2005incorporating,mann2010generalized,reichart2012multi,li2013joint,hu2016harnessing}. %These approaches, however, do not take advantage of the rich inter-dependencies in the input space.

Research capturing non-local dependencies in the input space have mostly relied on
%patterns (or templates: \cite{rusinol2013,bartoli2014}) and 
feature-based approaches. \citet{roberts2008extracting} and \citet{swampillai2011extracting} have designed intra- and inter-sentential features based on discourse and syntactic dependencies (e.g., shortest paths) to improve relation extraction. \citet{quirk2017distant} used document graphs to flexibly represent multiple types of relations between words (e.g., syntactic, adjacency and discourse relations).
% However, their work still relies on explicitly engineered sparse features.

Graph-based representations can be also learned with neural networks. The most related work to ours is the graph convolutional network by \citet{kipf2016semi}, which was developed to encode graph structures and perform node classification. In our framework, we adapt GCN as an intermediate module that learns non-local context, which --- instead of being used directly for classification --- is projected to the decoder to enrich local information and perform sequence tagging. 
%In this paper, we adapt GCN as an intermediate module to learn non-local context, and instead of directly classifying the nodes, we integrate their representations with the local encoder-decoder framework to perform sequence tagging. 
%In this paper, we adapt GCN to information extraction, designing it as an intermediate module to learn non-local context through graph convolutions between adjacent nodes. In our framework, the adapted GCN interacts with the encoder and the decoder, which are instead responsible to deal with the local information.
%In this paper, we adapt GCN to information extraction, designing it as an intermediate module to learn non-local context through graph convolutions between adjacent nodes, which are in turn encoded and decoded by other two modules, taking care of their local information.
%In this paper, we adapt GCN to work as an intermediary information extraction module that learn non-local context through graph convolutions between adjacent nodes, which are finally returned to the decoder to perform locally contexted tagging at the word level.

A handful of other information extraction approaches have used graph-based neural networks. \citet{miwa2016end} applied Tree LSTM \citep{tai2015improved} to jointly represent sequences and dependency trees for entity and relation extraction. On the same line of work, \citet{peng2017cross} and \citet{song2018n} introduced Graph LSTM, which extended the traditional LSTM to graphs by enabling a varied number of incoming edges at each memory cell. \citet{zhang2018graph} exploited graph convolutions to pool information over pruned dependency trees, outperforming existing sequence and dependency-based neural models in a relation extraction task.
These studies differ from ours in several respects.
%First, they propose models for relation extraction, whereas the present paper focuses on entity extraction. 
First, they can only model word-level graphs, whereas our framework can learn non-local context either from word- or sentence-level graphs, using it to reduce ambiguity during tagging at the word level. Second, all these studies achieved improvements only when using dependency trees. We extend the graph-based approach to validate the benefits of using other types of relations in a broader range of tasks, such as co-reference in named entity recognition, \textit{followed-by} link in social media, and layout structure in visual information extraction.

\begin{figure*}[t!]
\centering
\includegraphics[width=0.98\linewidth]{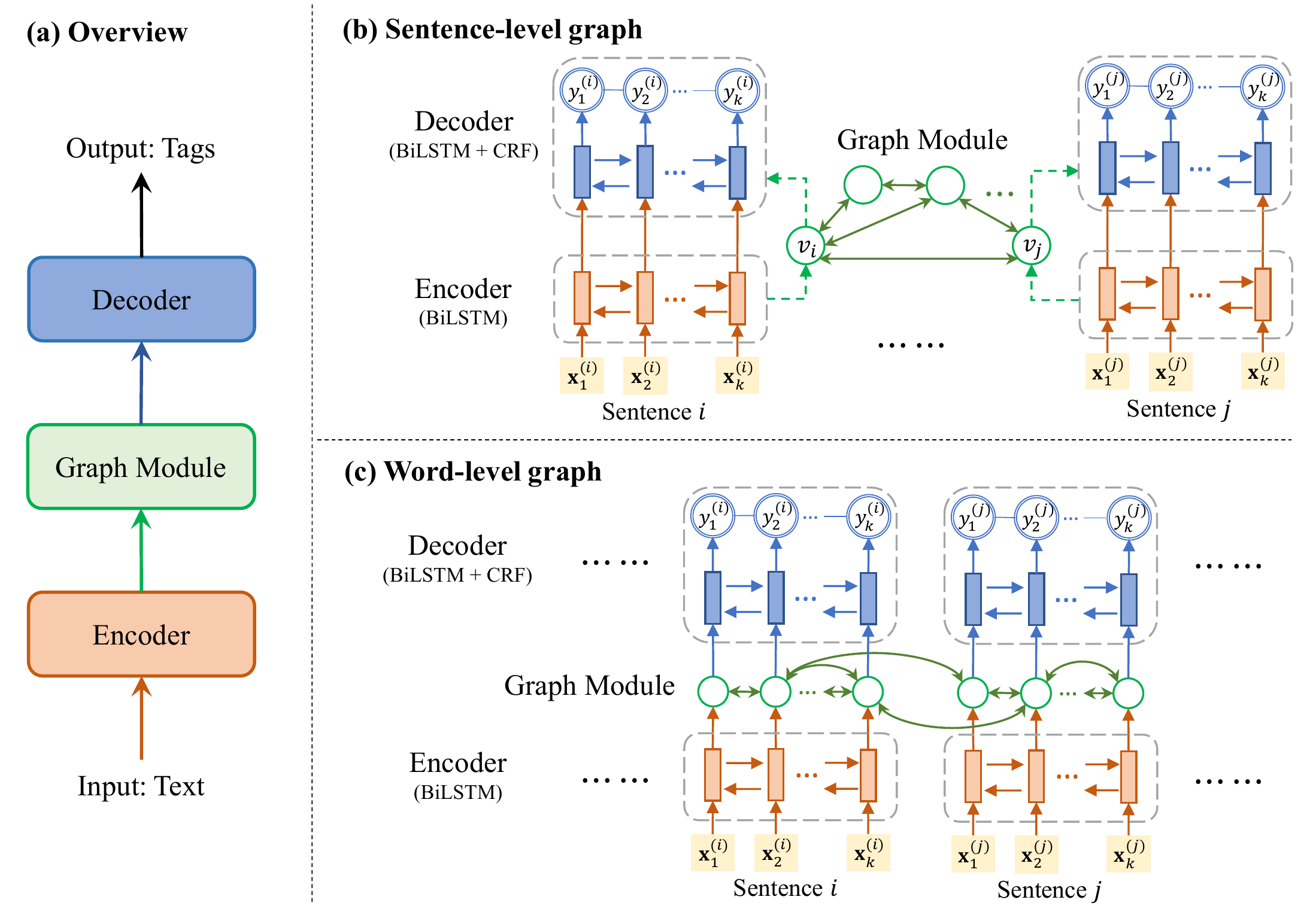}
% \vspace{-0.08in}
\caption{\label{fig:model}
GraphIE framework:
(a) an overview of the framework; 
(b) architecture for \textit{sentence-level graph}, where each sentence is encoded to a node vector and fed into the graph module, and the output of the graph module is used as the initial state of the decoder; 
(c) architecture for \textit{word-level graph}, where the hidden state for each word of the encoder is taken as the input node vector of the graph module, and then the output is fed into the decoder.}
% \vspace{-0.08in}
\end{figure*}

\section{Problem Definition}
We formalize information extraction as a sequence tagging problem. Rather than simply modeling inputs as sequences, we assume there exists a graph structure in the data that can be exploited to capture non-local and non-sequential dependencies between textual units, namely words or sentences.
% Our goal is therefore to develop a method which is able to take advantage of such dependencies to improve word-level information extraction.

We consider the input to be a set of sentences $S=\{s_1, \dots, s_N\}$ and an auxiliary graph $G=(V,E)$, where $V=\{v_1,\dots, v_M\}$ is the node set and $E\subset  V \times V$ is the edge set. Each sentence is a sequence of words. We consider two different designs of the graph: 
\begin{enumerate}[(1)]
    \item \textit{sentence-level graph}, where each node is a sentence (i.e. $M=N$), and the edges encode sentence dependencies; 
    \item \textit{word-level graph}, where each node is a word (i.e. $M$ is the number of words in the input), and the edges connect pairs of words, such as co-referent tokens. % or head-dependent pairs in syntactic trees. 
\end{enumerate}
The edges $e_{i,j}=(v_i, v_j)$ in the graph can be either directed or undirected. Multiple edge types can also be defined to capture different structural factors underlying the task-specific input data.
%  Assuming $\mathcal{T}_{\text{edge}}=\{\tau_1, \tau_2,\dots,\tau_K\}$ is the set of edge types and $K=|\mathcal{T}_{\text{edge}}|$, each edge has the type $\phi(e_{i,j})\in \mathcal{T}_{\text{edge}}$.

We use the BIO (Begin, Inside, Outside) tagging scheme in this paper. For each sentence $s_i= (w_1^{(i)}, w_2^{(i)}, \dots, w_k^{(i)})$,\footnote{While sentences may have different lengths, for notation simplicity we use a single variable $k$.} we sequentially tag each word as $\mathbf{y}_i = (y_1^{(i)}, y_2^{(i)}, \dots, y_k^{(i)})$. %  in the BIO format. % In this process, both local information from the textual unit and non-local information from the graph are taken into account. 

\section{Method}
\label{method}

GraphIE jointly learns local and non-local dependencies by iteratively propagating information between node representations. Our model has three components:

\begin{itemize}
    \item an \textit{encoder}, which generates local context-aware hidden representations for the textual unit (i.e. word or sentence, depending on the task) with a recurrent neural network;
    \item a \textit{graph module}, which captures the graph structure, learning non-local and non-sequential dependencies between textual units;
    \item a \textit{decoder}, which exploits the contextual information generated by the graph module to perform labelling at the word level.
\end{itemize}

Figure~\ref{fig:model} illustrates the overview of GraphIE and the model architectures for both sentence- and word-level graphs. In the following sections, we first introduce the case of the sentence-level graph, and then we explain how to adapt the model for the word-level graph.

\subsection{Encoder}
\label{encoder}
In GraphIE, we first use an encoder to generate text representations. Given a sentence $s_i=(w_1^{(i)}, w_2^{(i)}, \dots, w_k^{(i)})$ of length $k$,  each word $w_t^{(i)}$ is represented by a vector $\mathbf{x}_t^{(i)}$, which is the concatenation of its word embedding and a feature vector learned with a character-level convolutional neural network (CharCNN; \citet{kim2016character}). We encode the sentence with a recurrent neural network (RNN), defining it as 
\begin{equation}
\mathbf{h}_{1:k}^{(i)}
 = \mathtt{RNN}\left(\mathbf{x}_{1:k}^{(i)}~; \mathbf{0}, \Theta_\text{enc}\right),
\end{equation}
where $\mathbf{x}_{1:k}^{(i)}$ denotes the input sequence $[\mathbf{x}_1^{(i)},\cdots,\mathbf{x}_k^{(i)}]$, $\mathbf{h}_{1:k}^{(i)}$ denotes the hidden states $[\mathbf{h}_1^{(i)},\cdots,\mathbf{h}_k^{(i)}]$, $\mathbf{0}$ indicates the initial hidden state is zero, and $\Theta_\text{enc}$ represents the encoder parameters. We implement the RNN as a bi-directional LSTM \citep{hochreiter1997long}, and encode each sentence independently. 

We obtain the sentence representation for $s_i$ by averaging the hidden states of its words, i.e. 
    $\mathtt{Enc}(s_i) = \frac{1}{k} \left( \sum_{t=1}^k \mathbf{h}_t^{(i)} \right)$.
The sentence representations are then fed into the graph module.

% In some cases, we concatenate the positional encoding to the hidden vector  $\mathtt{Enc}(s_i)$ to inject information about the absolute or relative positions of the sentences \citep{zeng2014relation}.

\subsection{Graph Module}
\label{glob_structure}
The graph module is designed to learn the non-local and non-sequential information from the graph. 
We adapt the graph convolutional network (GCN) to model the graph context for information extraction.

Given the sentence-level graph $G = (V,E)$, where each node $v_i$ (i.e. sentence $s_i$) has the encoding $\mathtt{Enc}(s_i)$ capturing its local information, the graph module enriches such representation with neighbor information derived from the graph structure.

Our graph module is a GCN which takes as input the sentence representation, i.e. $\mathbf{g}_{i}^{(0)}=\mathtt{Enc}(s_i)$, and conducts graph convolution on every node, propagating information between its neighbors, and integrating such information into a new hidden representation. Specifically, each layer of GCN has two parts. The first gets the information of each node from the previous layer, i.e.
\begin{equation}
    \boldsymbol{\alpha}_{i}^{(l)} = \mathbf{W}_v^{(l)}\ \mathbf{g}_{i}^{(l-1)},
\end{equation}
where $\mathbf{W}_v^{(l)}$ is the weight to be learned.
The second aggregates information from the neighbors of each node, i.e. for node $v_i$, we have
\begin{equation}
\label{gcn_neighhbor}
\boldsymbol{\beta}_{i}^{(l)}  = \frac{1}{d(v_i)} \cdot \mathbf{W}_e^{(l)}    
\Bigg (\sum_{e_{i,j} \in E} \mathbf{g}_{j}^{(l-1)} \Bigg) ,
\end{equation}
where $d(v_i)$ is the degree of node $v_i$ (i.e. the number of edges connected to $v_i$) and is used to normalize $\boldsymbol{\beta}_{i}^{(l)}$, ensuring that nodes with different degrees have representations of the same scale.\footnote{We choose this simple normalization strategy instead of the two-sided normalization in \citet{kipf2016semi}, as it performs better in the experiments. The same strategy is also adopted by \citet{zhang2018graph}.}
In the simplest case, where the edges in the graph are undirected and have the same type, we use the same weight $\mathbf{W}_e^{(l)}$ for all of them. In a more general case, where multiple edge types exist, we expect them to have different impacts on the aggregation. Thus, we model these edge types with different weights in Eq.~\ref{gcn_neighhbor}, similar to the relational GCN proposed by \citet{schlichtkrull2018modeling}. When edges are directed, i.e. edge $e_{i,j}$ is different from $e_{j,i}$, the propagation mechanism should mirror such difference. In this case, we consider directed edges as two types of edges (forward and backward), and use different weights for them.

Finally, $\boldsymbol{\alpha}_{i}^{(l)}$  and $\boldsymbol{\beta}_{i}^{(l)}$ are combined to obtain the representation at the $l$-th layer,
\begin{equation}
\label{gcn}
\mathbf{g}_{i}^{(l)} =  \sigma \left( \boldsymbol{\alpha}_{i}^{(l)} + \boldsymbol{\beta}_{i}^{(l)} +b^{(l)}  \right),
\end{equation}
where $\sigma(\cdot)$ is the non-linear activation function, and $b^{(l)}$ is a bias parameter. 
% and $\lambda$ is the factor to balance local and neighbor information.
% Similar parameterization idea is also used in \cite{peng2017cross}.  

Because each layer only propagates information between directly connected nodes, we can stack multiple graph convolutional layers to get a larger receptive field, i.e. each node can be aware of more distant neighbors. After $L$ layers, for each node $v_i$ we obtain a contextual representation, 
% \begin{equation}
$\mathtt{GCN}(s_i)= \mathbf{g}_{i}^{(L)}$,
% \end{equation}
that captures both local and non-local information.

% GCN aggregates the neighbors' information with a simple average operation (i.e. Eq.~\ref{gcn_neighhbor}). \cite{velivckovic2017graph} proposed Graph Attention Networks (GAT) to improve the aggregation with attention mechanism. In our graph module, we can alternatively use GAT instead of GCN.

% We consider TGCN a powerful extension of GCN, as it handles different edge types with flexible parametrization. Similar idea is also used in the work of \cite{peng2017cross}, whose limitations are described in Section~\ref{related_work}. In Section~\ref{results}, we show that TGCN has stronger modeling power than GCN.

\subsection{Decoder}
\label{decoder}
To support tagging, the learned representation is propagated to the decoder.

In our work, the decoder is instantiated as a \text{BiLSTM+CRF} tagger \citep{lample-EtAl:2016:N16-1}. The output representation of the graph module, $\mathtt{GCN}(s_i)$, is split into two vectors of the same length, which are used as the initial hidden states for the forward and backward LSTMs, respectively.
% \footnote{Variants, such as concatenating the GCN-enriched sentence representation to each word's hidden state of the encoder, achieved lower performance.} 
In this way, the graph contextual information is propagated to each word through the LSTM.
Specifically, we have
\begin{equation}
\mathbf{z}_{1:k}^{(i)} 
 = \mathtt{RNN}\left(\mathbf{h}_{1:k}^{(i)}~; \mathtt{GCN}(s_i), \Theta_\text{dec}\right),
\label{eqn:decoder}
\end{equation}
where $\mathbf{h}_{1:k}^{(i)}$ are the output hidden states of the encoder, $\mathtt{GCN}(s_i)$ represents the initial state, and $\Theta_{\text{dec}}$ is the decoder parameters. A simpler way to incorporate the graph representation into the decoder is concatenating with its input, but the empirical performance is worse than using as the initial state.

Finally, we use a CRF layer \citep{lafferty2001conditional} on top of the \text{BiLSTM} to perform tagging,
\begin{equation}
    \mathbf{y}_i^* = \argmax_{\mathbf{y}\in \mathbf{Y}_k} p \left( \mathbf{y}\mid \mathbf{z}_{1:k}^{(i)}~; \Theta_\text{crf} \right),
    % p\big(y_t^{(i)}\mid v_i, G\big) = \text{softmax} \big( \mathbf{W}^y \ \mathbf{z}_t^{(i)} \big),
\end{equation}
where $\mathbf{Y}_k$ is the set of all possible tag sequences of length $k$, and $\Theta_\text{crf}$ represents the CRF parameters, i.e. transition scores of tags. CRF combines the local predictions of BiLSTM and the transition scores to model the joint probability of the tag sequence.\footnote{In GraphIE, the graph module models the input space structure, i.e. the dependencies between textual units (i.e. sentences or words), and the final CRF layer models the sequential connections of the output tags. Even though loops may exist in the input graph, CRF operates sequentially, thus the inference is tractable.}
%Although loops exist in the input graph, we use graph convolution to encode the graph structure and the CRF operates sequentially on the tag sequence. Thus we can still use the Viterbi algorithm for the inference of CRF.}

\subsection{Adaptation to Word-level Graphs}
GraphIE can be easily adapted to model word-level graphs. In such case, the nodes represent words in the input, i.e. the number of nodes $M$ equals the total number of words in the $N$ sentences. At this point, each word's hidden state in the encoder can be used as the input node vector $\mathbf{g}_i^{(0)}$ of the graph module. GCN can then conduct graph convolution on the word-level graph and generate graph-contextualized representations for the words. Finally, the decoder directly operates on the GCN's outputs, i.e. we change the BiLSTM decoder to
\begin{equation*}
    \mathbf{z}_{1:k}^{(i)}
    = \mathtt{RNN}\left(\left[\mathtt{GCN}(w_1^{(i)}),\cdots,\mathtt{GCN}(w_k^{(i)})\right]; \mathbf{0}, \Theta_\text{dec}\right),
\end{equation*}
where $\mathtt{GCN}(w_t^{(i)})$ is the GCN output for word $w_t^{(i)}$. In this case, the BiLSTM initial states are set to the default zero vectors. The CRF layer remains unchanged.

As it can be seen in Figure~\ref{fig:model}(c), the word-level graph module differs from the sentence-level one because it directly takes the word representations from the encoder and feeds its output to the decoder. In sentence-level graph, the GCN operates on sentence representations, which are then used as the initial states of the decoder BiLSTM.

%\begin{figure}[t]
%\centering
%\includegraphics[width=2.9in]{figs/doc_graph.jpg}
%\caption{\label{fig:doc_graph}Example: document graph of a report (Blue boxes represent nodes, and green lines represent edges.)}
%\end{figure}

\begin{table*}[t]
    \centering
    \setlength{\tabcolsep}{8pt}
    \small
    \begin{tabular}{llll}
    \toprule
    \bf Evaluation Task & \bf Graph Type & \bf Node    & \bf Edge  \\
    \midrule
    Textual IE        & word-level     & word       & \specialcell{1. non-local consistency (identical mentions)\\2. local sentential forward and backward} \\\midrule
    Social Media IE   & sentence-level & user's tweets       & \textit{followed-by} \\\midrule
    Visual IE         & sentence-level & text box & spatial layout (horizontal and vertical)\\    
    \bottomrule
    \end{tabular}
    % \vspace{-0.08in}
    \caption{Comparisons of graph structure in the three IE tasks used for evaluation.}
    \label{tab:eval-tasks}
\end{table*}

\section{Experimental Setup}
\label{evaluation}

%%%CHANGED
We evaluate the model on three tasks, including two traditional IE tasks, namely textual information extraction and social media information extraction, and an under-explored task --- \textit{visual information extraction}. For each of these tasks, we created a simple task-specific graph topology, designed to easily capture the underlying structure of the input data without any major processing.
% In the first two tasks, we utilize sentence-level graphs, and in the third task we utilize word-level graphs.
Table~\ref{tab:eval-tasks} summarizes the three tasks.

% \paragraph{GraphIE with GCN}
% In order to validate the benefits of using TGCN, we implement GraphIE also with GCN. Such variation ignores different edge types, using the same weights for all of them in the graph convolution.
% In Task 2, this model does not need to be reported because it only relies on one type of edge, and therefore TGCN and GCN would be the same.

\subsection{Task 1: Textual Information Extraction}
In this task, we focus on named entity recognition at discourse level (DiscNER). In contrast to traditional sentence-level NER (SentNER), where sentences are processed independently, in DiscNER, long-range dependencies and constraints across sentences have a crucial role in the tagging process. For instance, multiple mentions of the same entity are expected to be tagged consistently in the same discourse. Here we propose to use this (soft) constraint to improve entity extraction.

\paragraph{Dataset}
We conduct experiments on two NER datasets: the CoNLL-2003 dataset (\textsc{CoNLL03}) \citep{tjong2003introduction}, and the \textsc{Chemdner} dataset for chemical entity extraction \citep{krallinger2015chemdner}. We follow the standard split of each corpora. Statistics are shown in Table \ref{tab:ner-corpus}.

\paragraph{Graph Construction}
In this task, we use a word-level graph where nodes represent words. We create two types of edges for each document:
\begin{itemize}
    \item \textit{Local edges}: forward and backward edges are created between neighboring words in each sentence, allowing local contextual information to be utilized.
    \item \textit{Non-local edges}: re-occurrences of the same token other than stop words are connected, so that information can be propagated through, encouraging global consistency of tagging.\footnote{Note that other non-local relations such as co-references (cf. the example in Figure \ref{fig:example}) may be used for further improvement. However, these relations require additional resources to obtain, and we leave them to future work.}
\end{itemize}
%To build the non-local edges
%, for \textsc{Chemdner}, we exploit a named entity dictionary \citep{hettne2009dictionary}, which contains a noisy collection of 1.5M terms. For \textsc{CoNLL03}, 
%we consider all reoccurring nouns as potential mentions without introducing any additional resources.
%to ensure a fair comparison with previous work.

% A straightforward way to utilize the consistency constraint is to create edges between all pairs of the same potential entity mentions within a document. We explore two ways to approximate this graph (1) creating edges between every pair of nouns (2) 

\begin{table}[t]
    \centering
    \small
    \setlength{\tabcolsep}{5pt}
    \begin{tabular}{clccc}
    \toprule
    \multicolumn{2}{c}{\textsc{Dataset}}  & Train & Dev   & Test  \\
    \midrule
    \multirow{2}{*}{\textsc{CoNLL03}}  &  \#doc  & 946 & 216 & 231 \\
                              &  \#sent & 14,987 & 3,466 & 3,684 \\
    \cmidrule(lr){1-5}
    \multirow{2}{*}{\textsc{Chemdner}} &  \#doc  & 3,500 & 3,500 & 3,000 \\
                              &  \#sent & 30,739 & 30,796 & 26,399 \\
    \bottomrule
    \end{tabular}
    % \vspace{-0.08in}
    \caption{Statistics of the \textsc{CoNLL03} and the \textsc{Chemdner} datasets (Task 1).}
    \label{tab:ner-corpus}
% \end{minipage}
\end{table}

\subsection{Task 2: Social Media Information Extraction}
Social media information extraction refers to the task of extracting information from users' posts in online social networks \citep{benson2011event,li2014weakly}. In this paper, we aim at extracting \textit{education} and \textit{job} information from users' tweets. Given a set of tweets posted by a user, the goal is to extract mentions of the organizations to which they belong. The fact that the tweets are short, highly contextualized and show special linguistic features makes this task particularly challenging.

\paragraph{Dataset} We construct two datasets, \textsc{Education} and \textsc{Job}, from the Twitter corpus released by \citet{li2014weakly}. The original corpus contains millions of tweets generated by $\approx 10$ thousand users, where the \textit{education} and \textit{job} mentions are annotated using distant supervision  \citep{mintz2009distant}. We sample the tweets from each user, maintaining the ratio between positive and negative posts.\footnote{Positive and negative refer here to whether or not the \textit{education} or \textit{job} mention is present in the tweet.} % In \citet{li2014weakly}, sampling was not necessary because they processed the tweets and extracted engineered features beforehand, and then learned the model. 
The obtained \textsc{Education} dataset consists of $443,476$ tweets generated by $7,208$ users, and the \textsc{Job} dataset contains $176,043$ tweets generated by $1,772$ users. Dataset statistics are reported in \text{Table~\ref{tweet_entities}}.

The datasets are both split in 60\% for training, 20\% for development, and 20\% for testing. We perform 5 different random splits and report the average results.

\begin{table}[t]
\centering
\setlength{\tabcolsep}{6pt}
\small
\begin{tabular}{ccc}
\toprule
                       & \textsc{Education}      & \multicolumn{1}{c}{\textsc{Job}} \\ \midrule 
Users                  & 7,208          & 1,772           \\ %\hline
Edges                  & 11,167         & 3,498            \\% \hline
Positive Tweets        & 49,793         & 3,694           \\ %\hline
Negative Tweets        & 393,683        & 172,349       \\ 
\bottomrule
\end{tabular}
% \vspace{-0.08in}
\caption{Statistics of the \textsc{Education} and \textsc{Job} datasets (Task 2).}
\label{tweet_entities}
\end{table}

% Given the large amount of tweets in the original dataset, we performed sampling, \footnote{Positive and negative refer here to whether or not the \textit{education} or \textit{job} mention is present in the tweet. In \cite{li2014weakly} sampling was not necessary because they processed the tweets and extracted engineered features beforehand, and then learned the model.} We obtain two datasets of $443,476$ and $176,033$ tweets for the \textsc{Education} and \textsc{Job} respectively. Statistics about the sampled dataset are reported in \text{Table~\ref{tweet_entities}}.

\paragraph{Graph Construction} %We followed the preprocessing setup in \cite{li2014weakly}, where 
We construct the graph as \textit{ego-networks} \citep{leskovec2012learning}, i.e. when we extract information about one user, we consider the subgraph formed by the user and his/her direct neighbors. Each node corresponds to a Twitter user, who is represented by the set of posted tweets.\footnote{As each node is a set of tweets posted by the user, we encode every tweet with the encoder, and then average them to obtain the node representation. In the decoding phase, the graph module's output is fed to the decoder for each tweet.} Edges are defined by the \textit{followed-by} link, under the assumption that connected users are more likely to come from the same university or company. An example of the social media graph is reported in the appendices. %Only one type of edge is used in this task, and it is already provided in the original corpus.

%, already provided in the dataset. Only one type of edge is used here. The \textit{homophily} principle \cite{kwak2010twitter} suggests that connected users have higher probability to have the same education/job.

%\paragraph{Baseline\label{baseline}}
% We use the same SeqIE baseline as in Task 1, which takes in input one tweet at a time and tags its words. 

% For comparison, we also report the scores obtained by \cite{li2014weakly} in the entity-level classification task. The authors used a two-stage approach -- i.e. they performed Named Entity Recognition (NER) followed by candidate classification. In our model, we derive the entity representation by averaging the word representation in the last layer of GraphIE (Eq.~\ref{eqn:decoder}) and apply softmax for classification.\footnote{\label{note:compare}These results are reported only for reference, as \cite{li2014weakly} did not provide specific train/test split. On top of it, we have sampled the tweets.}

\subsection{Task 3: Visual Information Extraction}
Visual information extraction refers to the extraction of attribute values from documents formatted in various layouts. Examples include invoices and forms, whose format can be exploited to infer valuable information to support extraction.

\paragraph{Dataset} The corpus consists of 25,200 Adverse Event Case Reports (AECR) recording drug-related side effects. 
Each case contains an average of 9 pages. Since these documents are produced by multiple organizations, they exhibit large variability in the layout and presentation styles (e.g. text, table, etc.).\footnote{This dataset cannot be shared for patient privacy and proprietary issues. } The collection is provided with a separate human-extracted ground truth database that is used as a source of distant supervision.
%This poses an extra challenge in the task, as the human-extracted results do not always perfectly match the document text.\footnote{In order to reduce variability, the adopted matching rule ignores punctuation and case differences.}

Our goal is to extract eight attributes related to the patient, the event, the drug and the reporter (cf. Table~\ref{Tab:AECR} for the full list). Attribute types include dates, words and phrases --- which can be directly extracted from the document.

The dataset is split in 50\% cases for training, 10\% for development, and 40\% for testing. %In an extra experiment, we build a synthetic dataset to test GraphIE ability to generalize to unseen layouts.

\paragraph{Graph Construction} We first turn the PDFs to text using PDFMiner,\footnote{\url{https://euske.github.io/pdfminer/}} which provides words along with their positions in the page (i.e. bounding-box coordinates). Consecutive words are then geometrically joined into \textit{text boxes}. Each text box is considered as a ``sentence'' in this task, and corresponds to a \textit{node} in the graph.
%This process is notoriously error prone, as the matching rule might over- or under-tag. We expect however that the model handles such noise by observing a large number of samples.

Since the page layout is the major structural factor in these documents, we work on page-by-page basis, i.e. each page corresponds to a graph. The \textit{edges} are defined to horizontally or vertically connect \textit{nodes} (text boxes) that are close to each other (i.e. when the overlap of their bounding boxes, in either the vertical or horizontal direction, is over 50\%). Four types of edge are considered: left-to-right, right-to-left, up-to-down, and down-to-up. When multiple nodes are aligned, only the closest ones are connected. An example of visual document graph is reported in the appendices. %Figure~\ref{fig:mockup} shows two mock-up forms with textual units bounded by blue boxes and edges drawn as green lines. % Note that the graph is directed and symmetric, e.g., a left-to-right edge and the corresponding right-to-left edge must appear simultaneously.

% \paragraph{Evaluation Metrics} We compute entity-level prediction accuracy using precision, recall and F1 scores. Scores are calculated counting full matches between the postprocessed predictions\footnote{We applied two postprocessing steps to both GraphIE and the baseline: i) format consistency, which excludes outputs that do not present the expected attribute format (e.g., digits, dates or words); and ii) majority vote, which keeps the most frequent output for attributes that are known to have only one answer (e.g., patient's name).} and the human-extracted ground truths.

\subsection{Baseline and Our Method}
%%%CHANGED
%%%MAYBE TO BE FURTHER REINFORCED?
We implement a two-layer BiLSTM with a conditional random fields (CRF) tagger as the sequential baseline (SeqIE). This architecture and its variants have been extensively studied and demonstrated to be successful in previous work on information extraction \citep{lample-EtAl:2016:N16-1,ma-hovy:2016:P16-1}. In the textual IE task (Task 1), our baseline is shown to obtain competitive results with the state-of-the-art method in the \textsc{CONLL03} dataset. In the visual IE task (Task 3), in order to further increase the competitiveness of the baseline, we sequentially concatenate the horizontally aligned text boxes, therefore fully modeling the horizontal edges of the graph.

%%%CHANGED
Our baseline shares the same encoder and decoder architecture with GraphIE, but without the graph module. Both architectures have similar computational cost. In Task 1, we apply GraphIE with word-level graph module (cf. Figure~\ref{fig:model}(c)), and in Task 2 and Task 3, we apply GraphIE with sentence-level graph module (cf. Figure~\ref{fig:model}(b)). 
%The graph module can be implemented in either GCN or its extension with attention mechanism (i.e. GAT). However, we do not observe improvement with GAT in the sentence-level cases. 

\subsection{Implementation Details}
The models are trained with $\mathtt{Adam}$ \citep{kingma2014adam} to minimize the CRF objective. For regularization, we choose dropout with a ratio of 0.1 on both the input word representation and the hidden layer of the decoder. The learning rate is set to 0.001. We use the development set for early-stopping and the selection of the best performing hyperparameters. For CharCNN, we use 64-dimensional character embeddings and 64 filters of width 2 to 4 \citep{kim2016character}. The 100-dimensional pretrained GloVe word embeddings \citep{pennington2014glove} are used in Task 1 and 2, and 64-dimensional randomly initialized word embeddings are used in Task 3. We use a two-layer GCN in Task 1, and a one-layer GCN in Task 2 and Task 3. The encoder and decoder BiLSTMs have the same dimension as the graph convolution layer. In Task 3, we concatenate a positional encoding to each text box's representation by transforming its bounding box coordinates to a vector of length 32, and then applying a $\tanh$ activation.

%%% TURNING IT INTO ENGLISH
\section{Results}
\label{results}

\begin{table}[t]
    \centering
    \small
    \begin{tabular}{cll}
    \toprule
    \textsc{Dataset} & Model & F1 \\
    \midrule
    \multirow{5}{*}{\textsc{CoNLL03}}
                             & \citet{lample-EtAl:2016:N16-1} & 90.94 \\
                             & \citet{ma-hovy:2016:P16-1} & 91.21 \\
                             & \citet{ye2018hybrid} & 91.38 \\
                        % \cmidrule(lr){2-3}
                             & SeqIE & 91.16 \\
                             & GraphIE & \bf{91.74}$^*$ \\
%                              & GraphIE (GAT) & \bf 91.34 \\

    \cmidrule(lr){1-3}
    \multirow{3}{*}{\textsc{Chemdner}}
                              & \citet{krallinger2015chemdner} & 87.39 \\
                        % \cmidrule(lr){2-3}
                              & SeqIE & 88.28 \\
                              %& SeqIE+dict & 82.49 \\
                              & GraphIE & \bf 89.71$^*$ \\
%                               & GraphIE (GAT) & \bf 85.70 \\
    \bottomrule
    \end{tabular}
    % \vspace{-0.08in}
    \caption{NER accuracy on the \textsc{CoNLL03} and the \textsc{Chemdner} datasets (Task 1). Scores for our methods are the average of 5 runs. *~indicates statistical significance of the improvement over SeqIE ($p<0.01$).} % Results from \citet{lample-EtAl:2016:N16-1} and \citet{ma-hovy:2016:P16-1} are listed here as references, as they utilize similar architectures to SeqIE, but enhanced with more task-specific engineering.}
    \label{tab:ner-results}
\end{table}

\begin{figure}[t]
    \centering
    \hspace{-0.15in}
    \includegraphics[height=1.2in]{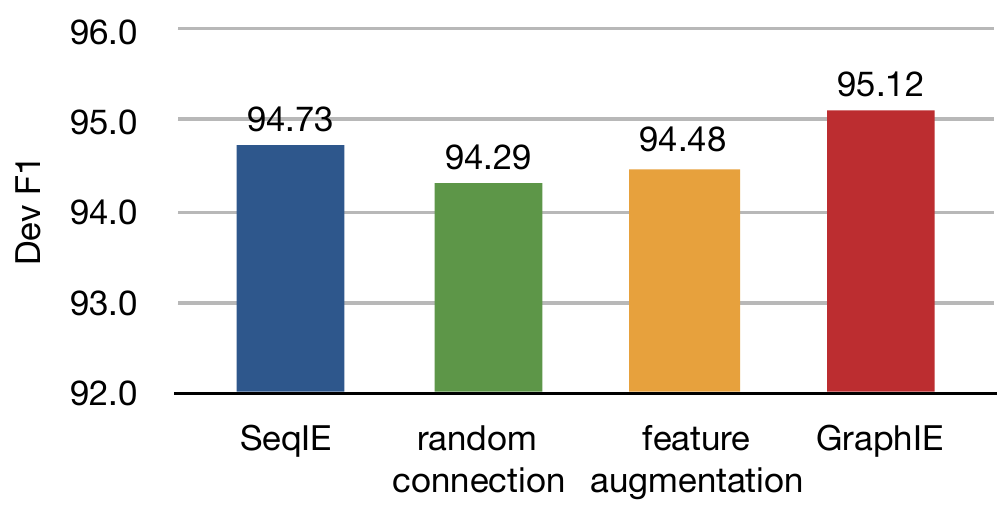}
    \caption{Analysis on the \textsc{CoNLL03} dataset. We compare with two alternative designs: (1) \textit{random connection}, where we replace the constructed graph by a random graph with the same number of edges; (2) \textit{feature augmentation}, where we use the average embedding of each node and its neighbors as the input to the decoder, instead of the GCN which has additional parameters. We report F1 scores on the development set.}
    \label{fig:conll}
\end{figure}

\begin{table*}[t]
\centering
\small
\begin{tabular}{cccccccccl}
\toprule
\multirow{2}{*}{\textsc{Dataset}} & \multicolumn{3}{c}{Dictionary} & \multicolumn{3}{c}{SeqIE} & \multicolumn{3}{c}{GraphIE}  \\ \cmidrule(lr){2-4} \cmidrule(lr){5-7} \cmidrule(lr){8-10} 
&  P   &   R   &  F1  &  P   &   R   &  F1  &   P   &  R    &   \multicolumn{1}{c}{F1}   \\ \midrule
{\textsc{Education}} 
& 78.7 & 93.5 & 85.4 & 85.2 & 93.6 & 89.2 & 92.9 & 92.8 & \textbf{92.9}$^*$  \\
{\textsc{Job}}         
& 55.7 & 70.2 & 62.1 & 66.2 & 66.7 & 66.2 & 67.1 & 66.1 & \textbf{66.5}  \\ 
\bottomrule 
\end{tabular}
% \vspace{-0.08in}
\caption{Extraction accuracy on the \textsc{Education} and \textsc{Job} datasets (Task 2). Dictionary is a naive method which creates a dictionary of entities from the training set and extracts their mentions during testing time. Scores are the average of 5 runs. * indicates the improvement over \text{SeqIE} is statistically significant (Welch's $t$-test, $p<0.01$).}
\label{Tab:Twitter_Entity}
\end{table*}

\subsection{Task 1: Textual Information Extraction}
Table~\ref{tab:ner-results} describes the NER accuracy on the \textsc{CoNLL03} \citep{tjong2003introduction} and the \textsc{Chemdner} \citep{krallinger2015chemdner} datasets.

For \textsc{CoNLL03}, we list the performance of existing approaches.
% \footnote{The performance of existing approaches for \textsc{Chemdner} is not reported because the authors used different settings from ours, so that results are not directly comparable.} 
Our baseline SeqIE obtains competitive scores compared to the best methods. The fact that GraphIE significantly outperforms it, highlights once more the importance of modeling non-local and non-sequential dependencies and confirms that our approach is an appropriate method to achieve this goal.\footnote{We achieve the best reported performance among methods not using the recently introduced ELMo \cite{peters2018deep} and BERT \cite{devlin2018bert}, which are pretrained on extra-large corpora and computationally demanding.}

%For both datasets, we list the performance of existing approaches, including the state-of-the-art methods without external resources \cite{ye2018hybrid,krallinger2015chemdner}.
% \footnote{The performance of existing approaches for \textsc{Chemdner} is not reported because the authors used different settings from ours, so that results are not directly comparable.} 
%As it can be seen, our baseline SeqIE already obtains competitive scores compared to the existing best approaches. The fact that GraphIE significantly outperforms it on both datasets, establishing new state-of-the-art performance, highlights once more the importance of modeling non-local and non-sequential dependencies and confirms that our approach is an appropriate method to achieve this goal.

For \textsc{Chemdner}, we show the best performance reported in \citet{krallinger2015chemdner}, obtained with a feature-based method. Our baseline outperforms the feature-based method, and GraphIE further improves the performance by $1.4\%$.

\paragraph{Analysis}
To understand the advantage of GraphIE, we first investigate the importance of graph structure to the model. As shown in Figure~\ref{fig:conll}, using random connections clearly hurts the performance, bringing down the F1 score of GraphIE from 95.12\% to 94.29\%. It indicates that the task-specific graph structures introduce beneficial inductive bias. Trivial feature augmentation also does not work well, confirming the necessity of learning the graph embedding with GCN. 

We further conduct error analysis on the test set to validate our motivation that GraphIE resolves tagging ambiguity by encouraging consistency among identical entity mentions (cf. Figure \ref{fig:example}). Here we examine the word-level tagging accuracy. We define the words that have more than one possible tags in the dataset as \textit{ambiguous}. We find that among the $1.78\%$ tagging errors of SeqIE, $1.16\%$ are \textit{ambiguous} and $0.62\%$ are \textit{unambiguous}. GraphIE reduces the error rate to $1.67\%$, with $1.06\%$ to be \textit{ambiguous} and $0.61\%$ \textit{unambiguous}. We can see that most of the error reduction indeed attributes to the \textit{ambiguous} words. %, indicating a higher tagging consistency with GraphIE.

%For both benchmarks, we report the performance of GraphIE using either GCN or GAT. 
%For the \textsc{Chemdner} dataset, we also report the performance of the enhanced baseline (SeqIE+dict), built by concatenating to the decoder hidden states a Boolean feature indicating whether the current token exists in a dictionary.

%Our baseline SeqIE achieves comparative performance as the existing best approaches. GraphIE significantly outperforms SeqIE on both datasets, and achieves new state-of-the art performance on the \textsc{CoNLL03} dataset. % and also outperforms SeqIE+dict on the \textsc{Chemdner} dataset, 
%It demonstrates that (1) non-local information is beneficial for discourse-level entity extraction and that (2) GraphIE offers a coherent framework for effectively exploiting such non-local dependencies.

% With respect to the difference in performance between GAT and GCN, we conjecture that...

%well-designed relational structures are beneficial for discourse-level entity extraction and (2) GraphIE offers a coherent way for exploiting these relational constraints.

\subsection{Task 2: Social Media Information Extraction} 
%\paragraph{Performance}
Table~\ref{Tab:Twitter_Entity} shows the results for the social media information extraction task. We first report a simple dictionary-based method as a baseline. Neural IE models achieve much better performance, showing that meaningful patterns are learned by the models rather than simply remembering the entities in the training set. The proposed \text{GraphIE} outperforms SeqIE in both the \textsc{Education} and \textsc{Job} datasets, and the improvements are more significant for the \textsc{Education} dataset ($3.7\%$ versus $0.3\%$). The reason for such difference is the variance in the affinity scores \citep{mislove2010you} between the two datasets. \citet{li2014weakly} underline that affinity value for \textsc{Education} is $74.3$ while for \textsc{Job} it is only $14.5$, which means that in the datasets neighbors are $5$ times more likely to have studied in the same university than worked in the same company. We can therefore expect that a model like GraphIE, which exploits neighbors' information, obtains larger advantages in a dataset characterized by higher affinity. 

% We also observe that both models perform better on \textsc{Education} than \textsc{Job}. The reason might be that educational institutions can be recognized by affixes or suffixes (e.g., \textit{University of}), while this is unlikely for other organizations.

\subsection{Task 3: Visual Information Extraction}
Table~\ref{Tab:AECR} shows the results in the visual information extraction task. GraphIE outperforms the SeqIE baseline in most attributes, and achieves $1.2\%$ improvement in the mirco average F1 score. It confirms that the benefits of using layout graph structure in visual information extraction.

\begin{table}[t]
\begin{adjustbox}{center}
\small
\setlength{\tabcolsep}{4pt}
\begin{tabular}{ccccccl}
\toprule
\multirow{2}{*}{\textsc{Attribute}}   & \multicolumn{3}{c}{SeqIE}  & \multicolumn{3}{c}{GraphIE} \\ \cmidrule(lr){2-4} \cmidrule(lr){5-7} 
                                         & P & R & F1 & P & R & \multicolumn{1}{c}{F1}  \\ 
\midrule
\multicolumn{1}{l}{\textit{P. Initials}}        & 93.5 & 92.4 & \textbf{92.9} 
                                        & 93.6 & 91.9 & 92.8 \\ %\hline
\multicolumn{1}{l}{\textit{P. Age}}             & 94.0 & 91.6 & 92.8 
                                        & 94.8 & 91.1 & \textbf{92.9} \\ %\hline
\multicolumn{1}{l}{\textit{P. Birthday}}   & 96.6 & 96.0 & \textbf{96.3} 
                                        & 96.9 & 94.7 & 95.8 \\ %\hline
\multicolumn{1}{l}{\textit{Drug Name}}           & 71.2 & 51.2 & 59.4 
                                        & 78.5 & 50.4 & \textbf{61.4} \\ %\hline
\multicolumn{1}{l}{\textit{Event}}               & 62.6 & 65.2 & 63.9 
                                        & 64.1 & 68.7 & \textbf{66.3} \\ %\hline
\multicolumn{1}{l}{\textit{R. First Name}}      & 78.3 & 95.7 & 86.1 
                                        & 79.5 & 95.9 & \textbf{86.9} \\ %\hline
\multicolumn{1}{l}{\textit{R. Last Name}}       & 84.5 & 68.4 & 75.6 
                                        & 85.6 & 68.2 & \textbf{75.9} \\ %\hline
\multicolumn{1}{l}{\textit{R. City}}            & 88.9 & 65.4 & 75.4
                                        & 92.1 & 66.3 & \textbf{77.1} \\ %\hline
\midrule
\multicolumn{1}{l}{Avg. (macro)}       & 83.7 & 78.2 & 80.3 
                                        & 85.7 & 78.4 & \textbf{81.1}$^\dagger$   \\
\multicolumn{1}{l}{Avg. (micro)}       & 78.5 & 73.8 & 76.1 
                                        & 80.3 & 74.6 & \textbf{77.3}$^\dagger$   \\
\bottomrule
\end{tabular}
\end{adjustbox}
\caption{Extraction accuracy on the AECR dataset (Task 3). Scores are the average of 5 runs. \textit{P.} is the abbreviation for \textit{Patient}, and \textit{R.} for \textit{Reporter}. $\dagger$ indicates statistical significance of the improvement over SeqIE ($p<0.05$).}
\label{Tab:AECR}
\end{table}

% The average F1 score of the GraphIE w/ TGCN is $69.7\%$, which is $+5.4$ points over SeqIE. SeqIE can perform reasonably well because several attributes appear sequentially in the document, similarly to what happens in the second example in Figure~\ref{fig:mockup}. The two variants of GraphIE show however higher modelling skills, achieving consistent improvements on all the attributes. With respect to the edge types, then, GraphIE w/TGCN outperforms GraphIE w/GCN by $+0.6$ points, which confirms the benefits of modeling different edge types.

The extraction performance varies across the attributes, ranging from $61.4\%$ for \textit{Drug Name} to $95.8\%$ for \textit{Patient Birthday} (similar variations are visible in the baseline). Similarly, the gap between GraphIE and SeqIE varies in relation to the attributes, ranging between $-0.5\%$ in \textit{Patient Birthday} and $2.4\%$ in \textit{Event}.  %Such a wide variance can be ascribed to two factors: attribute type and attribute ambiguity. With respect to the attribute type, it is harder to fully extract long phrases (e.g. \textit{Reported Event} and \textit{Drug Name}), thus the reported scores for these attributes are lower. For what concerns attribute ambiguity, all systems perform better on (semi)closed-class attributes (e.g. \textit{Reporter City} which contains mostly U.S. main cities) and on attributes that have less viable candidates in the same document (e.g. \textit{Patient Date of Birth}, \textit{Patient Age} and \textit{Patient Initial}). These attributes are less ambiguous and therefore relatively easier to identify.

In the ablation test described in Table~\ref{Tab:Ablation}, we can see the contribution of: using separate weights for different edge types ($+0.8\%$), horizontal edges ($+3.1\%$), vertical edges ($+5.4\%$), and CRF ($+5.7\%$). 

% Also the gap between GraphIE and SeqIE varies across the attributes. It is lower for attributes where the field name and its value are often organized in a sequential way (e.g. \textit{Patient Date of Birth}), and higher for less linear ones (e.g. \textit{Reporter City}). On top of it, GraphIE achieves higher improvements on highly ambiguous attributes, such as \textit{Reporter First Name} and \textit{Reporter Last Name}. We suggest that this higher disambiguation ability is obtained by the access to a wider amount of non-local context.

\begin{table}[t]
\centering
\small
\begin{tabular}{ll}
\toprule
Model                         & Dev F1 \\
\midrule
GraphIE                       & 77.8  \\
\quad -- Edge types           & 77.0\ ($\downarrow$ 0.8) \\
\quad -- Horizontal edges     & 74.7\ ($\downarrow$ 3.1) \\
\quad -- Vertical edges       & 72.4\ ($\downarrow$ 5.4) \\
\quad -- CRF                  & 72.1\ ($\downarrow$ 5.7) \\
\bottomrule
\end{tabular}
% \vspace{-0.08in}
\caption{Ablation study (Task 3). Scores are micro average F1 on the development set. ``--'' means removing the element from GraphIE.}
\label{Tab:Ablation}
\end{table}

\begin{figure}[t]
\centering
\hspace{-0.15in}
\includegraphics[width=0.3\textwidth]{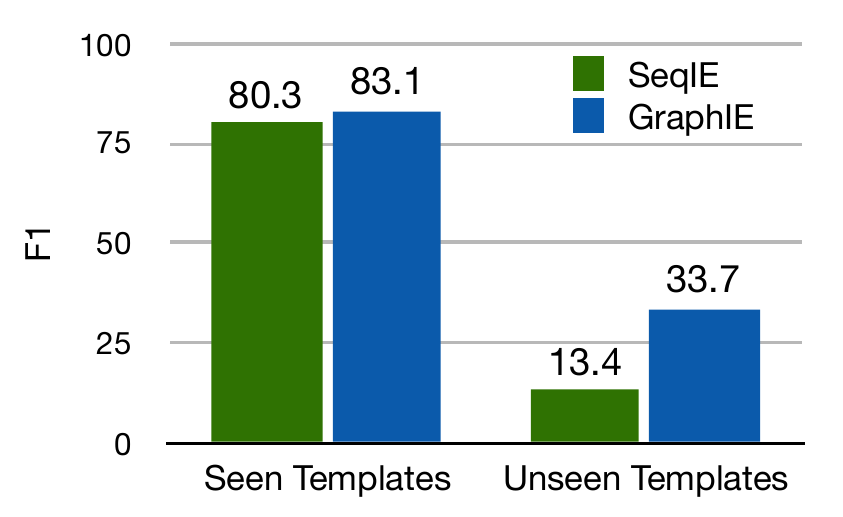}
% \vspace{-0.08in}
\caption{Micro average F1 scores tested on \textit{seen} and \textit{unseen} templates (Task 3).}
\label{fig:generalization}
\end{figure}

%%% SIGNIFICANCE OF THE GAP RATHER THAN PERFORMANCE
\paragraph{Generalization}% After showing the disambiguation ability of our model, 
We also assess \text{GraphIE}'s capacity of dealing with unseen layouts through an extra analysis. From our dataset, we sample $2,000$ reports containing the three most frequent templates, and train the models on this subset. Then we test all models in two settings: 1) \textit{seen templates}, consisting of $1,000$ additional reports in the same templates used for training; and 2) \textit{unseen templates}, consisting of $1,000$ reports in two new template types.

The performance of GraphIE and SeqIE is reported in Figure~\ref{fig:generalization}. Both models achieve good results on \textit{seen templates}, with GraphIE still scoring $2.8\%$ higher than SeqIE. The gap becomes even larger when our model and the sequential one are tested on \textit{unseen templates} (i.e. $20.3\%$), demonstrating that by explicitly modeling the richer structural relations, GraphIE achieves better generalizability.

\section{Conclusions}
\label{conclusions}
%%%CHANGED
We introduced GraphIE, an information extraction framework that learns local and non-local contextual representations from graph structures to improve predictions. The system operates over a task-specific graph topology describing the underlying structure of the input data. GraphIE jointly models the node (i.e. textual units, namely words or sentences) representations and their dependencies. Graph convolutions project information through neighboring nodes to finally support the decoder during tagging at the word level.

%%%CHANGED
We evaluated our framework on three IE tasks, namely textual, social media and visual information extraction. Results show that it efficiently models non-local and non-sequential context, consistently enhancing accuracy and outperforming the competitive SeqIE baseline (i.e. \text{BiLSTM+CRF}).

%%%CHANGED
Future work includes the exploration of automatically learning the underlying graphical structure of the input data.

\section*{Acknowledgments}
We thank the MIT NLP group and the reviewers for their helpful comments. This work is supported by MIT-IBM Watson AI Lab. Any opinions, findings, conclusions, or recommendations expressed in this paper are those of the authors, and do not necessarily reflect the views of the funding organizations.

\bibliography{reference}
\bibliographystyle{acl_natbib}

\clearpage
\appendix
\setcounter{page}{1}

\section{Examples of the Constructed Graphs}
\subsection{Social Media Information Extraction}
\setcounter{figure}{0}
\begin{figure}[h]
\centering
\includegraphics[width=.42\textwidth]{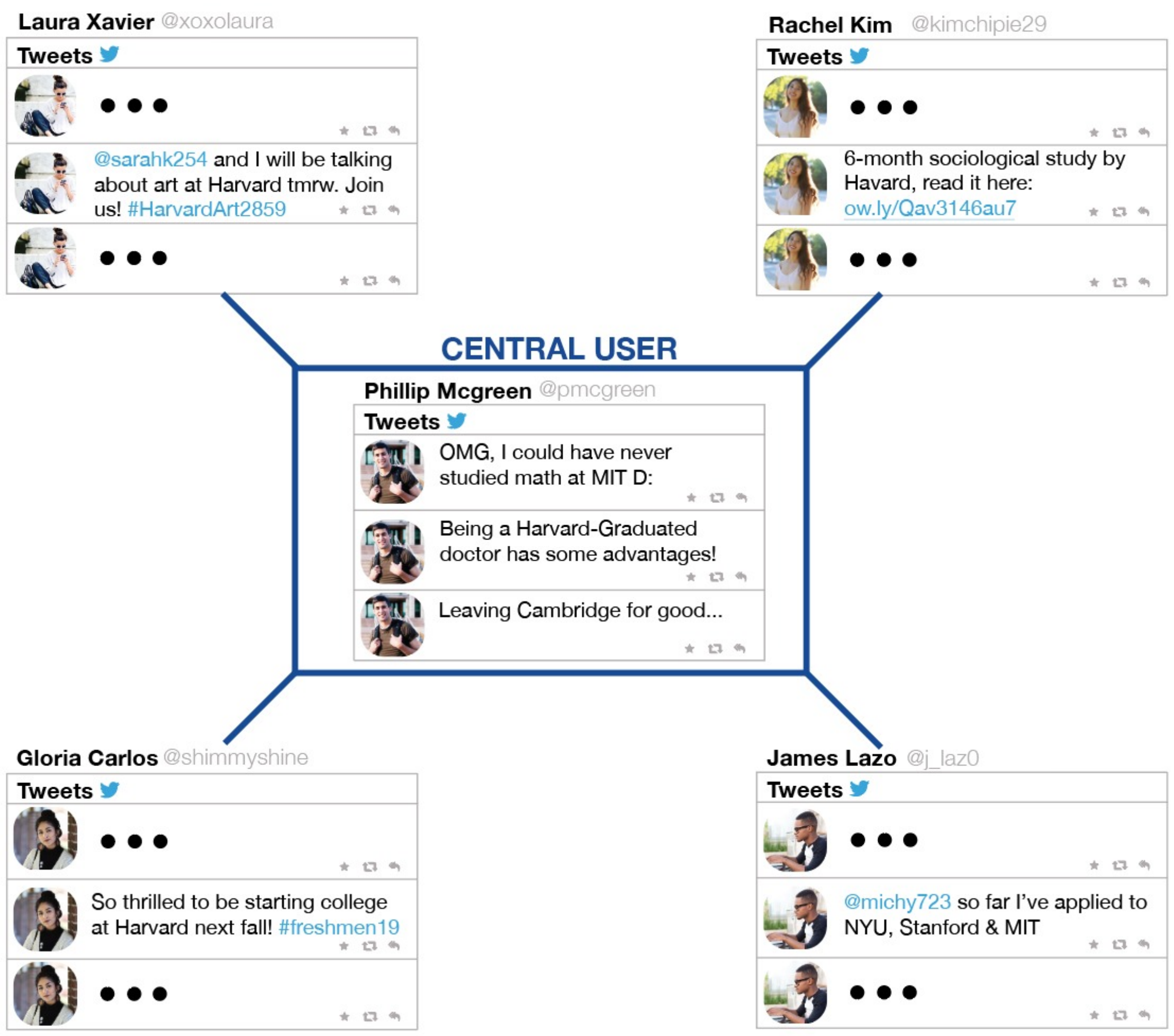}
\caption{Mock-up example of \textsc{Education} Social Media Information Extraction (Task 1). Nodes are represented as users and edges are \textit{follow-by} relations.}
\label{fig:mockup1}
\end{figure}

% \newpage

\subsection{Visual Information Extraction}
\begin{figure}[h]
\centering
\includegraphics[width=.47\textwidth]{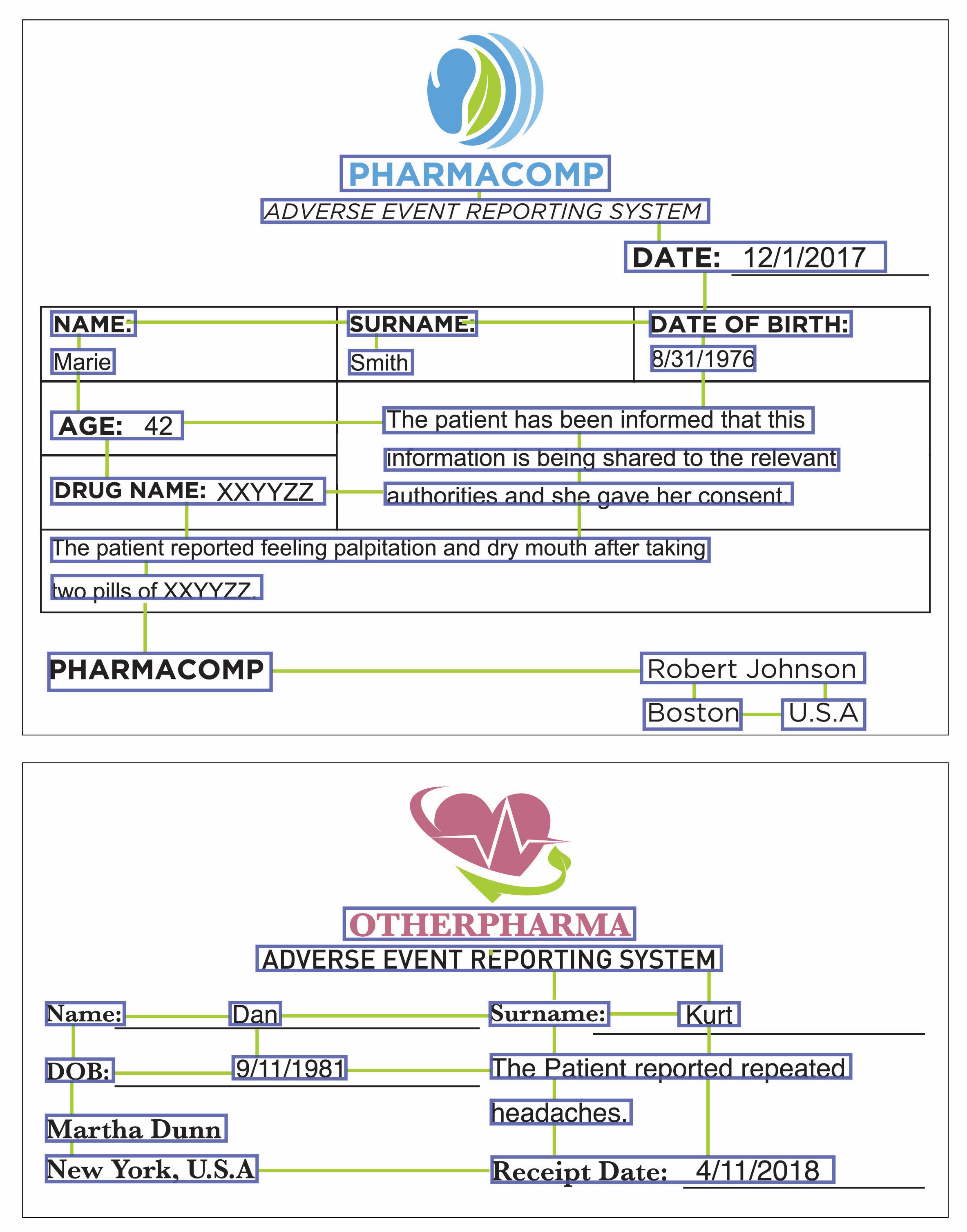}
\vspace{-0.1in}
\caption{Mock-up example of two forms with different layouts (Task 3). Graphical dependencies are shown as green lines connecting text in blue bounding-boxes.}
\label{fig:mockup2}
\end{figure}

\end{document}